\documentclass[conference]{IEEEtran}
\IEEEoverridecommandlockouts

\usepackage{cite}
\usepackage{amsmath,amssymb,amsfonts}
\usepackage{algorithmic}
\usepackage{graphicx}
\usepackage{textcomp}
\usepackage{xcolor}
\usepackage{multirow}
\usepackage{float}
\usepackage{hyperref}
\usepackage{afterpage}

\def\BibTeX{{\rm B\kern-.05em{\sc i\kern-.025em b}\kern-.08em
    T\kern-.1667em\lower.7ex\hbox{E}\kern-.125emX}}
\begin{document}

\title{Improving LLM-Powered EDA Assistants with RAFT}
\author{
\IEEEauthorblockN{Luyao Shi}
\IEEEauthorblockA{\textit{IBM Research} \\
San Jose, CA \\
luyao.shi@ibm.com}
\and
\IEEEauthorblockN{Michael Kazda}
\IEEEauthorblockA{\textit{IBM Infrastructure} \\
Poughkeepsie, NY \\
kazda@us.ibm.com}
\and
\IEEEauthorblockN{Charles Schmitter}
\IEEEauthorblockA{\textit{IBM Infrastructure} \\
Poughkeepsie, NY \\
charles.schmitter@ibm.com}
\and
\IEEEauthorblockN{Hemlata Gupta}
\IEEEauthorblockA{\textit{IBM Infrastructure} \\
Poughkeepsie, NY \\
guptah@us.ibm.com}
}



\maketitle

\begin{abstract}
Electronic design engineers often struggle to efficiently access relevant information for tasks like design verification and technology development. While large language models (LLMs) can enhance productivity as conversational agents, pre-trained open-source LLMs lack domain-specific knowledge for Electronic Design Automation (EDA). In a Retrieval-Augmented Generation (RAG) context, LLMs rely on external context but may still produce inaccurate responses. Retrieval-Augmented Fine-Tuning (RAFT) improves LLM performance, but acquiring labeled question/answer (Q/A) data in EDA is difficult. To address this, we propose using synthetic Q/A datasets to enhance LLMs with RAFT. Our results show that RAFT with synthetic data significantly boosts LLM performance for RAG-based EDA tasks. We also investigate the impact of using real user questions as Retrieval-Augmented Few-Shot (RAFS) examples for synthetic data generation. Additionally, we implement secure access control to ensure sensitive information is only accessible to authorized personnel. Finally, we assess the risk of data leakage and unintended memorization during fine-tuning with synthetic data, providing practical insights.
\end{abstract}


\section{Introduction}
Design engineers often face substantial challenges in navigating complex workflows, particularly in design construction and verification. A major hurdle in large organizations is efficiently accessing relevant documentation or identifying the right subject-matter expert for guidance. Documentation may exist in multiple, redundant versions or be scattered across different locations, making retrieval cumbersome. These difficulties are especially pronounced for new employees or when integrating unfamiliar tools, where specialized terminology, jargon, and acronyms create additional barriers. An intelligent, always-available assistant capable of providing timely and accurate information could greatly enhance engineering efficiency and decision-making.

Large language models (LLMs) \cite{achiam2023gpt} have become powerful tools for generating human-like responses across diverse tasks. However, their reliability is limited by outdated training data and the absence of domain-specific or proprietary knowledge. Additionally, LLMs often hallucinate, confidently producing incorrect responses when faced with unfamiliar topics—an issue particularly problematic in engineering applications where precision is essential. Retrieval-Augmented Generation (RAG) \cite{lewis2020retrieval} addresses this challenge by integrating external knowledge retrieval into response generation, ensuring outputs are accurate and contextually relevant. Notable applications in design environments include ChipNemo \cite{liu2023chipnemo}, a domain-specific LLM for EDA that leverages RAG to enhance accuracy and reduce hallucinations, and Ask-EDA \cite{shi2024ask}, which combines hybrid RAG with Abbreviation De-Hallucination (ADH) to provide more accurate and contextually relevant responses.

Ensuring data security is a critical concern when utilizing LLMs, particularly in specialized fields such as chip design, where proprietary information could be exposed through third-party API interactions. By developing domain-specific in-house models, organizations can mitigate these risks while also achieving superior performance on specialized tasks. Research has shown that tailoring LLMs for a particular domain not only enhances security but also improves accuracy and reliability. Recent efforts, such as ChipNemo \cite{liu2023chipnemo}, ChatEDA \cite{wu2024chateda}, and OpenROAD-Assistant \cite{sharma2024openroad}, have demonstrated the benefits of fine-tuning LLMs with EDA-specific data, leading to more effective results in the field.

Domain-specific fine-tuning improves LLM performance but faces challenges such as outdated knowledge, high data requirements, and overfitting. Although RAG can address these limitations by allowing models to retrieve external information instead of relying solely on static parameters, it has its own drawbacks. A general-purpose LLM may struggle to integrate retrieved domain-specific information coherently, leading to inconsistencies or misinterpretations in its responses. Additionally, RAG depends on the quality and relevance of retrieved documents, and retrieval failures can still result in hallucinations. Retrieval-Augmented Fine-Tuning (RAFT) \cite{zhang2024raft} mitigates these issues by optimizing models specifically for RAG. RAFT fine-tunes the LLM to better integrate retrieved context into its generation process, adapting the model to dynamically retrieve and condition its responses on real-time data from a knowledge base. While RAFT has shown promising results in the EDA domain \cite{sharma2024openroad}, it still relies on labeled question-answer pairs \cite{wu2024eda}, which are often scarce or unavailable in many technical fields.

In this paper, we propose using synthetic question-answer datasets to enhance LLMs with RAFT, demonstrating significant improvements for RAG-based EDA assistants. Synthetic data enables the utilization of broadly available unlabeled documents spanning various technical domains, reducing reliance on scarce manually labeled datasets. We also explore the impact of incorporating real user questions as Retrieval-Augmented Few-Shot (RAFS) examples in synthetic data generation. Additionally, we implement secure access control in our EDA-LLM assistant, ensuring sensitive information is accessible only to authorized personnel through retrieval-based context management. To assess potential risks, we investigate whether Retrieval-Augmented Fine-Tuning (RAFT) on synthetic data leads to unintended memorization and data leaks, providing practical considerations based on our findings.

We highlight our key contributions as follows:
\begin{itemize}
    \item We propose using synthetic question-answer datasets to enhance LLMs with RAFT, leveraging broadly available unlabeled documents to reduce reliance on scarce labeled data and improve performance for RAG-based EDA assistants.
    \item We propose incorporating real user questions as Retrieval-Augmented Few-Shot (RAFS) examples in synthetic data generation, assessing its benefits for improving LLM performance.
    \item We implement a retrieval-based access control mechanism, ensuring that sensitive information remains accessible only to authorized personnel while maintaining usability.
    \item We examine whether RAFT fine-tuning on synthetic data leads to unintended memorization and data leakage, offering practical considerations to mitigate these risks.
\end{itemize}


\begin{figure*}[!h]
    \centering
    \includegraphics[width=0.84\textwidth]{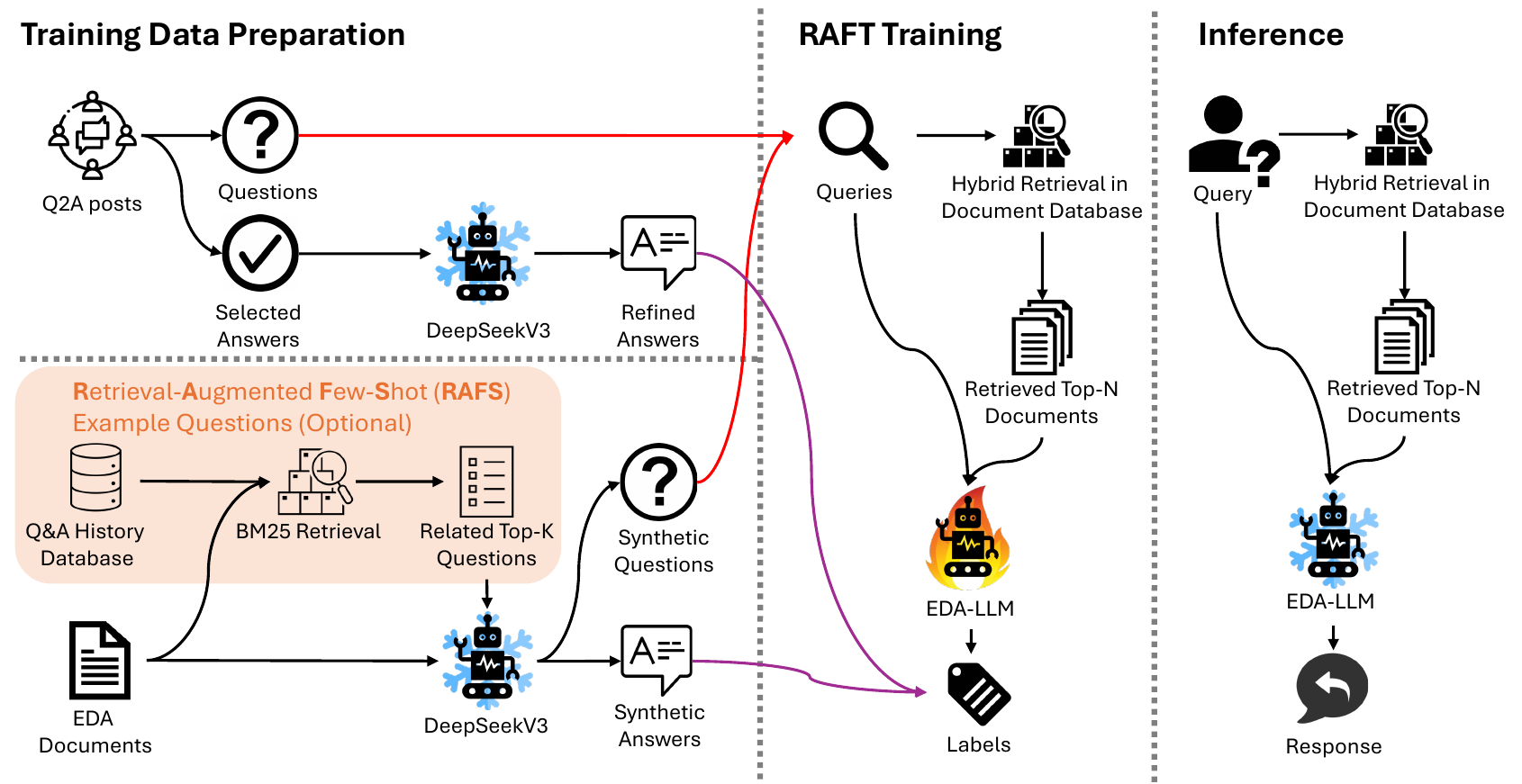}
    \caption{Overview of the workflow for training data preparation, RAFT training, and inference.}
    \label{fig:flow}
\end{figure*}

\section{Methods}
\label{sec:methods}
The fine-tuning workflow for LLM-powered EDA assistants (EDA-LLM) is shown in Figure \ref{fig:flow}. 

\subsection{Training Data Preparation}
We utilized two sources of data and prepared two types of training sets for fine-tuning EDA-LLM.

\begin{figure}[h]
    \centering
    \includegraphics[width=0.47\textwidth]{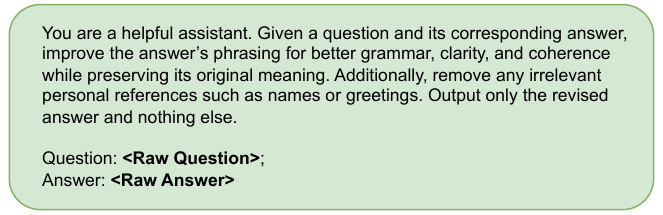}
    \caption{Prompt used for refining answers in Q2A posts for better grammar, clarity, and coherence.}
    \label{fig:prompt2}
\end{figure}

\textbf{Q2A posts:} We have a Stack Overflow-style system where engineers can pose questions, and subject matter experts (SMEs) provide answers and comments. A voting system allows the best answer to be marked. These expert responses form a collective knowledge base with ground-truth labeled data. However, human-written answers often follow a free-form style and include unnecessary elements like greetings (e.g., ``Hi John, to answer your question..."). To enhance clarity, grammar, and coherence, we use DeepSeek-V3 to refine these answers. The refined question-answer dataset is incorporated into the training set to improve the EDA-LLM's response structure and clarity. The prompt used for refinement is shown in Figure \ref{fig:prompt2}.

\begin{figure}[h]
    \centering
    \includegraphics[width=0.47\textwidth]{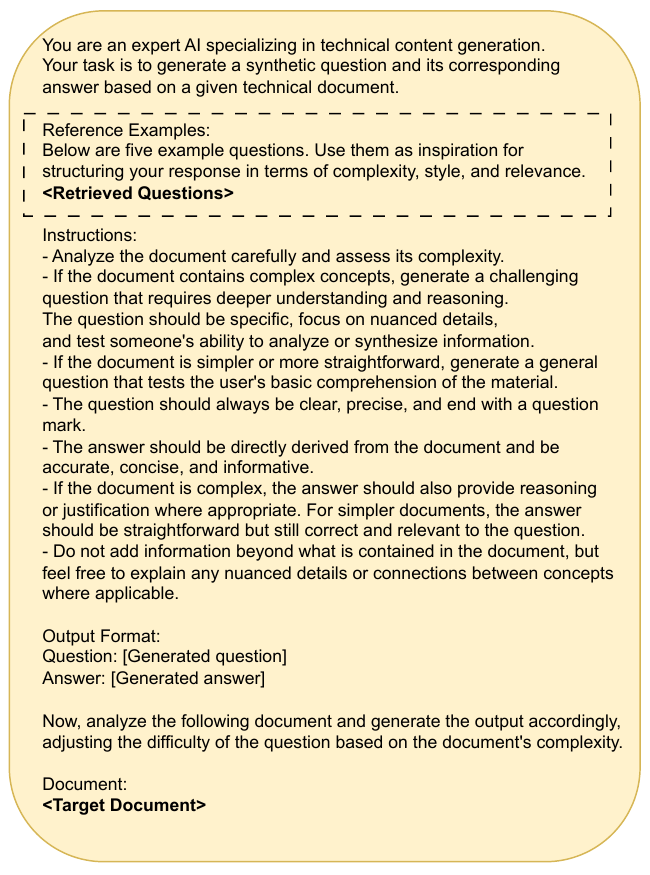}
    \caption{Prompt for generating synthetic question-answer pairs from a target document. The five retrieval-augmented example questions used as few-shot examples (highlighted in the dashed box) are optional.}
    \label{fig:prompt1}
\end{figure}

\begin{figure}[h]
    \centering
    \includegraphics[width=0.47\textwidth]{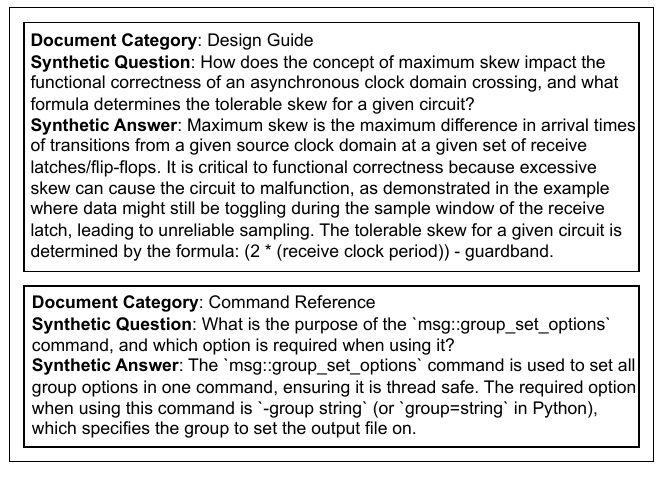}
    \caption{Examples of synthetic Q\&A pairs. Questions from Design Guides are more complex, requiring deeper reasoning, while those from Command References are more straightforward.}
    \label{fig:exampleQA}
\end{figure}

\textbf{Synthetic QA:} Although Q2A posts provide expert responses, their quantity is often limited. Furthermore, when training EDA-LLM with RAFT, there is no guarantee that the relevant document will appear in the retrieved context. To address this, we leverage a larger pool of unlabeled EDA documents and employ DeepSeek-V3 to generate synthetic questions and corresponding answers based on these documents. The prompt used for generating synthetic question-answer pairs from a given document is shown in Figure \ref{fig:prompt1}. We observed that certain documents, such as Design Guides, are more complex than others, like Command References. Therefore, in the prompt we encourage LLM to first analyze the document and assess its complexity: if the document contains complex concepts, we ask LLM to generate a challenging question that requires deeper understanding and reasoning; if the document is simpler or more straightforward, we ask LLM to generate a general question that tests the user's basic comprehension of the material. We provide a couple of synthetic Q\&A examples in Figure \ref{fig:exampleQA}. As observed, the synthetic question generated from a Design Guide document is more complex, requiring deeper understanding and reasoning, while the question based on a Command Reference document is more straightforward. In both cases, the synthetic answer effectively addresses the question.

We also maintain a collection of past user questions and corresponding LLM-generated responses from the deployment of the EDA assistant. These questions can serve as Retrieval-Augmented Few-Shot (RAFS) examples, guiding the Q\&A-generation LLM (DeepSeek-V3 here) to produce questions that more closely resemble real human inquiries in terms of complexity and style. In our implementation, we first concatenate each question with its corresponding LLM response in the Q\&A History Database. This concatenation provides additional context for subsequent relevance searches. Then, given a target EDA document, we use BM25 search to identify the Top-K (K=5) most relevant concatenated Q\&A pairs. These Top-K questions are used as few-shot examples in the prompt (highlighted in the dashed box in Figure \ref{fig:prompt2}) for synthetic Q\&A generation (indicated in the orange-shaded section of Figure \ref{fig:flow}).

\subsection{Retrieval Augmented Fine-Tuning (RAFT)}
\begin{figure}[h]
    \centering
    \includegraphics[width=0.47\textwidth]{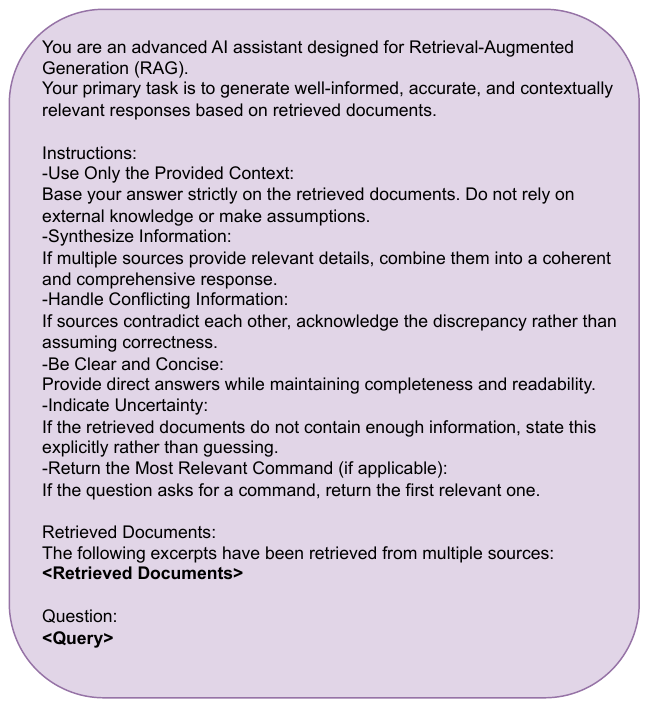}
    \caption{Prompt used for RAFT training and RAG-based inference.}
    \label{fig:prompt3}
\end{figure}

Both Q2A posts and the Synthetic QA dataset are used to fine-tune EDA-LLM with RAFT. As illustrated in Figure \ref{fig:flow}, given a query, we first perform a hybrid search to retrieve the Top-N relevant document chunks from our ingested document database. Following \cite{shi2024ask}, we employ Reciprocal Rank Fusion (RRF) to combine semantic search results (using sentence transformer \cite{reimers2019sentence} embeddings) with lexical search results (using BM25 \cite{robertson2009probabilistic}). The query and the Top-N retrieved document chunks are then incorporated into the prompt, training EDA-LLM to generate the ground truth answer. Similarly, at inference we keep the EDA-LLM frozen and ask EDA-LLM to generate response based on the provided query and retrieved context. The prompts used for RAFT training and RAG-based inference are shown in Figure \ref{fig:prompt3}.

\subsection{Secure Access Control}
A key consideration in building an LLM-based EDA assistant is implementing secure access control over ingested data. Some documents, such as restricted design guides, contain sensitive information and should only be accessible to authorized personnel, while general EDA documents, like command references, are available to everyone in the organization.

To enforce access control, we associate each document chunk with its corresponding access group during ingestion. When a user submits a query, we retrieve their access group information and use Elasticsearch \cite{elasticsearch} to filter only the documents they are authorized to access. These filtered documents are then incorporated into the prompt and sent to the LLM for response generation.

However, in this setting, it is crucial to ensure that EDA-LLM does not memorize sensitive information during RAFT, as this could lead to unintended data leaks to unauthorized personnel. To validate this, we created a test set, SynthQA\_MissingContext, derived from the Synthetic QA training set.

In this evaluation, we modify the prompt by removing any retrieved document chunks originating from the target EDA document used to generate the synthetic Q\&A. This prevents the model from relying on those sources when answering the question. Our goal is to determine whether the fine-tuned LLM has memorized information from training and can still produce answers despite the absence of relevant retrieved documents. Additionally, we explore whether incorporating some training samples with missing context can help the model appropriately refuse to generate a response when it lacks sufficient information.

\section{Evalaution}
\label{sec:eval}

\subsection{Datasets}
We collected 749 Q2A posts, as described in Section \ref{sec:methods}.A. We filtered out posts without a marked best answer and those where the answer consisted solely of a reference link (e.g., a webpage link). The remaining 405 Q\&A posts were refined using DeepSeek-V3 and then randomly split into a training set of 305 samples and a test set of 100 samples. We refer to this test set as Q2A throughout the paper.

For synthetic QA generation, we first removed extremely short EDA documents (fewer than 1,000 characters) and sampled 1,000 documents. Among these, longer documents were truncated to retain only the first 10,000 characters. The document categories are listed in Table \ref{tab:1}. Within each category, we aimed to split the samples using a 90\%/10\% ratio whenever the sample size allowed. The corresponding numbers are also shown in Table \ref{tab:1}. Following the steps in \ref{sec:methods}.A, we used DeepSeek-V3 to generate one synthetic Q\&A pair for each EDA document. We refer to this test set as SynthQA throughout the paper. For both Q2A and SynthQA questions, we used hybrid search with RRF to retrieve the Top-N (N=10) document chunks and included the query along with the Top-N retrieved chunks in the prompt for each sample. Each chunk is 2,000 characters long, with a 200-character overlap. Note that the retrieval space consists of our entire document collection, comprising approximately 200,000 chunks from 14,000 documents.

As outlined in Section \ref{sec:methods}.C, we further sampled 100 synthetic Q\&A pairs from the SynthQA training set and removed the relevant retrieved document chunks from the prompt to simulate missing context. We refer to these 100 test samples as SynthQA\_MissingContext throughout the paper.

\begin{table}[h]
\centering
\renewcommand{\arraystretch}{1.2}  
\caption{Document categories and their corresponding train/test splits for the synthetic QA dataset.}
\label{tab:1}
\begin{tabular}{|l|c|c|}
\hline
\textbf{Category/Number}     & \textbf{Train} & \textbf{Test} \\ \hline
Parameter Reference & 1     & 1    \\ \hline
Timing              & 24    & 3    \\ \hline
DevOps              & 237   & 26   \\ \hline
Design Guide        & 295   & 33   \\ \hline
Command Reference   & 343   & 37   \\ \hline
Total               & 900   & 100  \\ \hline
\end{tabular}
\end{table}

\subsection{Implementation Details}
 DeepSeek-V3 \cite{liu2024deepseek} is a strong Mixture-of-Experts (MoE) language model with 671B total parameters with 37B activated for each token. For training data preparation, we leverage DeepSeek-V3\footnote{\url{https://huggingface.co/deepseek-ai/DeepSeek-V3}} to generate high-quality refined answers and synthetic Q\&As, as illustrated in Figure \ref{fig:flow}. In our experiments, we use Llama-3.1-8B-Instruct\footnote{\url{https://huggingface.co/unsloth/Meta-Llama-3.1-8B-Instruct-bnb-4bit}}\cite{grattafiori2024llama} as our EDA-LLM assistant and fine-tune it. Additionally, we compare its performance with several baseline LLM models, including Mistral-7B-Instruct-v0.3\footnote{\url{https://huggingface.co/unsloth/mistral-7b-instruct-v0.3-bnb-4bit}}, Mistral-Nemo-Instruct-2407\footnote{\url{https://huggingface.co/unsloth/Mistral-Nemo-Instruct-2407}} (a 12B model), Mistral-Small-Instruct-2409\footnote{\url{https://huggingface.co/unsloth/Mistral-Small-Instruct-2409}} (a 22B model), and Llama-3.1-70B-Instruct\footnote{\url{https://huggingface.co/unsloth/Meta-Llama-3.1-70B-Instruct}}.

In our experiment, we use Unsloth \cite{unsloth} for fast training and low memory usage. The EDA-LLM models are fine-tuned with Parameter-Efficient Fine-Tuning (PEFT) and Low-Rank Adaptation (LoRA) \cite{hu2022lora}, applying low-rank adaptation with a LoRA rank of 128, $\alpha = 32$, and dropout set to 0. All models are trained with a maximum sequence length of 8,192 tokens, 4-bit quantization, and gradient checkpointing. Training is performed for 5 epochs with a batch size of 8, 2 gradient accumulation steps, a learning rate of $2 \times 10^{-5}$, a warmup ratio of 0.1, weight decay of 0, and a cosine learning rate scheduler. All trainings are conducted on a single Nvidia A100 80GB GPU. Note that for fair comparison, all baseline model inferences also use Unsloth with 4-bit quantization (except for DeepSeek-V3, which is used for high-quality Q\&A generation).

For hybrid retrieval, we use Reciprocal Rank Fusion (RRF) to combine semantic search results (from sentence transformer embeddings) with lexical search results (using BM25). Our semantic search is performed with our in-house bi-encoder sentence transformer model IBM Slate 125m \cite{ibmSlate125mV2}, although our method is agnostic to the choice of embedding model.

\subsection{Metrics}
BERTScore \cite{zhang2019bertscore} and BARTScore \cite{yuan2021bartscore} are evaluation metrics that assess the quality of generated text beyond exact word overlaps. BERTScore computes token-level embeddings from models like BERT to measure semantic similarity in a context-aware manner. Specifically, we use ``microsoft/deberta-xlarge-mnli" as the BERT scorer model, which is one of the best performers.

BARTScore measures the log-likelihood of one text sequence being generated given another text sequence. It evaluates the semantic and contextual alignment between a generated text and a reference text using the probabilities assigned by the BART model. Note that the raw scores of BARTScore are negative log probabilities that are difficult to explain. Therefore we use normalized BART scores for precision and recall calculated as follows\footnote{\url{https://huggingface.co/datasets/tau/scrolls/blame/7d4a9140e65653bf8bc2aece85c3b774cd1ff7c8/metrics/bart_score.py}}:

\begin{equation}
    \text{Precision} = \frac{\text{BARTScore}(\textit{Ref}, \textit{Ref})}{\text{BARTScore}(\textit{Ref}, \textit{Pred})}
\label{eq:precision}
\end{equation}

\begin{equation}
    \text{Recall} = \frac{\text{BARTScore}(\textit{Ref}, \textit{Ref})}{\text{BARTScore}(\textit{Pred}, \textit{Ref})}
\label{eq:recall}
\end{equation}

In these equations, \textit{Pred} refers to the predicted text, and \textit{Ref} refers to the reference text. As shown in Equations \ref{eq:precision} and \ref{eq:recall}, BARTScore-based precision and recall are normalized by the self-similarity of \textit{Ref}, so that the score range is between 0 and 1. Following \cite{yuan2021bartscore}, we also add prompt to guide the BART model’s understanding of the task. Specifically, we add the prompt \textit{z} before \textit{y} in $\text{BARTScore}(\textit{x}, \textit{y})$, where \textit{z} $\in \{$ ``That is to say, '', ``In other words, '', ``To rephrase it, '', ``i.e., '' $\}$, and report the averaged scores over the 4 prompts.
Finally, the F1 score was calculated as the arithmetic average of precision and recall. We used the ``facebook/bart-large-cnn'' model fine-tuned on ``ParaBank2'' as our BART scorer.

\begin{table*}[h]
\caption{The main evaluation results on the Q2A and SynthQA test sets. The highest score for each metric on each dataset is \textbf{bold}.}
\label{tab:2}
\centering
\renewcommand{\arraystretch}{1.2}  
\setlength{\tabcolsep}{4pt}  
\begin{tabular}{|l|cccccc|cccccc|}
\hline
\multicolumn{1}{|c|}{\multirow{3}{*}{\textbf{Models/Metrics}}} & \multicolumn{6}{c|}{\textbf{Q2A}}                                                                                        & \multicolumn{6}{c|}{\textbf{SynthQA}}                                                                                    \\ \cline{2-13} 
\multicolumn{1}{|c|}{}                                         & \multicolumn{3}{c|}{\textbf{BERT Score (\%)}}                         & \multicolumn{3}{c|}{\textbf{BART Score (\%)}}    & \multicolumn{3}{c|}{\textbf{BERT Score (\%)}}                         & \multicolumn{3}{c|}{\textbf{BART Score (\%)}}    \\ \cline{2-13} 
\multicolumn{1}{|c|}{}                                         & Precision      & Recall         & \multicolumn{1}{c|}{F1}             & Precision      & Recall         & F1             & Precision      & Recall         & \multicolumn{1}{c|}{F1}             & Precision      & Recall         & F1             \\ \hline
Mistral-7B-Instruct-v0.3                                       & 78.60          & 81.00          & \multicolumn{1}{c|}{79.66}          & 32.73          & 36.68          & 34.71          & 79.74          & 81.87          & \multicolumn{1}{c|}{80.72}          & 26.12          & 30.44          & 28.28          \\
Mistral-Nemo-Instruct-2407                                     & 77.55          & 79.23          & \multicolumn{1}{c|}{78.22}          & 29.51          & 34.22          & 31.86          & 75.03          & 80.09          & \multicolumn{1}{c|}{77.40}          & 21.27          & 29.52          & 25.39          \\
Mistral-Small-Instruct-2409                                    & 80.99          & 82.33          & \multicolumn{1}{c|}{81.56}          & 35.28          & 39.52          & 37.40          & 78.70          & 81.74          & \multicolumn{1}{c|}{80.11}          & 25.17          & 31.72          & 28.44          \\
Llama-3.1-8B-Instruct                                          & 79.61          & 80.28          & \multicolumn{1}{c|}{79.84}          & 32.73          & 36.73          & 34.73          & 75.73          & 80.13          & \multicolumn{1}{c|}{77.80}          & 21.06          & 29.89          & 25.48          \\
Llama-3.1-70B-Instruct                                         & 82.52          & 81.41          & \multicolumn{1}{c|}{81.85}          & 38.21          & 38.51          & 38.36          & 79.74          & 81.54          & \multicolumn{1}{c|}{80.56}          & 25.90          & 30.84          & 28.37          \\ \hline
EDA-Llama-8B-D1                                                & 84.28          & 84.38          & \multicolumn{1}{c|}{84.26}          & 41.40          & 42.50          & 41.95          & 78.25          & 79.93          & \multicolumn{1}{c|}{79.01}          & 24.13          & 27.90          & 26.02          \\
EDA-Llama-8B-D2                                                & 84.35          & 84.75          & \multicolumn{1}{c|}{84.47}          & 41.05          & 43.50          & 42.28          & \textbf{83.47} & 82.76          & \multicolumn{1}{c|}{\textbf{83.07}} & \textbf{31.45} & 31.48          & \textbf{31.47} \\
EDA-Llama-8B-D2-RAFS                                             & \textbf{84.95} & \textbf{85.52} & \multicolumn{1}{c|}{\textbf{85.17}} & \textbf{43.69} & \textbf{45.48} & \textbf{44.59} & 82.86          & \textbf{82.77} & \multicolumn{1}{c|}{82.78}          & 29.99          & \textbf{31.50} & 30.74          \\ \hline
\end{tabular}
\end{table*}

\subsection{Results}
The main evaluation results on the Q2A and SynthQA test sets are presented in Table \ref{tab:2}. For the fine-tuned EDA-Llama-8B model using RAFT, ``D1” refers to training on a single data source, the Q\&A posts, which contain 305 training samples. ``D2” denotes using two data sources by incorporating an additional 900 training samples from SynthQA (total 1,205 training samples). ``D2-RAFS” also uses two training data sources, but utilizes SynthQA with retrieval-augmented few-shot example questions (also 1,205 training samples). It is worth mentioning that the synthetic questions and answers from ``D2” and ``D2-RAFS” are different. Also note that the 100 SynthQA test examples are generated without RAFS question examples.

As shown in Table \ref{tab:2}, retrieval-augmented fine-tuning (RAFT) with EDA domain-specific data (EDA-Llama-8B-D1) significantly improves performance on the Q2A test set compared to the baseline LLM models. However, on the SynthQA test set, only precision improves, while recall does not. This suggests that training on a limited Q2A dataset helps the LLM generate more concise responses, but the lower recall may stem from the increased complexity and varied styles of SynthQA questions. Notably, Q2A questions are human-authored and often broad, whereas SynthQA consists of synthetic questions derived from specific documents, requiring deeper comprehension and reasoning about technical details.

After incorporating the SynthQA training data (D2), the model's performance further improves on both the Q2A and SynthQA test sets, with particularly significant gains on the SynthQA test results.

Introducing Retrieval Augmented Few-Shot (RAFS) question examples in the prompt for SynthQA training (D2-RAFS) further enhances performance on the Q2A test set but not on SynthQA. This may be because the few-shot examples are retrieved from a past user question database, which better aligns with the style of Q2A test questions. This finding suggests that for real-world EDA assistant deployment, where the model interacts with human users, incorporating RAFS question examples in the SynthQA training process could be more beneficial.

\begin{table*}[h]
\caption{Exploratory results on handling absent context in the RAG prompt. Lower recall on SynthQA\_MissingContex indicates less memorization when relevant context is missing, while recall should remain high on SynthQA when relevant context is present. A higher number of ``I don't know" (\#IDK) responses should occur on SynthQA\_MissingContext and fewer on SynthQA.}
\label{tab:3}
\centering
\renewcommand{\arraystretch}{1.2}  
\setlength{\tabcolsep}{4pt}  
\begin{tabular}{|l|cllcllc|cllcllc|}
\hline
\multicolumn{1}{|c|}{\multirow{2}{*}{\textbf{Models/Metrics}}} & \multicolumn{7}{c|}{\textbf{SynthQA\_MissingContext}}                                                               & \multicolumn{7}{c|}{\textbf{SynthQA}}                                                                           \\ \cline{2-15} 
\multicolumn{1}{|c|}{}                                         & \multicolumn{3}{c|}{\textbf{BERT Recall (\%)}} & \multicolumn{3}{c|}{\textbf{BART Recall (\%)}} & \textbf{\#IDK}    & \multicolumn{3}{c|}{\textbf{BERT Recall (\%)}} & \multicolumn{3}{c|}{\textbf{BART Score (\%)}} & \textbf{\#IDK} \\ \hline
Llama-3.1-8B-Instruct                                          & \multicolumn{3}{c|}{74.86}                     & \multicolumn{3}{c|}{20.73}                     & 6                 & \multicolumn{3}{c|}{80.13}                     & \multicolumn{3}{c|}{29.89}                    & 0              \\ \hline
EDA-Llama-8B-D2                                                & \multicolumn{3}{c|}{76.57}                     & \multicolumn{3}{c|}{21.01}                     & 0                 & \multicolumn{3}{c|}{82.76}                     & \multicolumn{3}{c|}{31.48}                    & 0              \\
EDA-Llama-8B-D3-MC10\%                                         & \multicolumn{3}{c|}{66.78}                     & \multicolumn{3}{c|}{16.72}                     & 64                & \multicolumn{3}{c|}{81.11}                     & \multicolumn{3}{c|}{29.72}                    & 8              \\
EDA-Llama-8B-D3-MC20\%                                         & \multicolumn{3}{c|}{65.84}                     & \multicolumn{3}{c|}{26.23}                     & 70                & \multicolumn{3}{c|}{79.67}                     & \multicolumn{3}{c|}{28.91}                    & 12             \\ \hline
\end{tabular}
\end{table*}

Table \ref{tab:3} presents exploratory results on handling absent context in the RAG prompt. Specifically, we evaluate the ``SynthQA\_MissingContext" test set (see Section \ref{sec:eval}.A) to assess the extent to which the fine-tuned model memorizes ground truth answers and generates responses even when no source context is provided. This is a critical consideration for implementing secure access control in RAG, ensuring the EDA-LLM assistant does not inadvertently disclose sensitive information to unauthorized users. Ideally, recall should be as low as possible on ``SynthQA\_MissingContext" while remaining high on SynthQA. However, compared to the baseline Llama-8B model, the fine-tuned D2 model exhibits increased recall, suggesting that RAFT fine-tuning leads the LLM to memorize answers to some extent, even when relevant documents are absent from the RAG context. It is also worth noting that the recall scores for the baseline model on ``SynthQA\_MissingContext" are not particularly low. This may be because, even without fine-tuning on EDA data, the baseline model attempts to extract relevant information from retrieved chunks and generate a response. We manually verified that the baseline model responded with an ``I don't know" type of answer (\#IDK in Table \ref{tab:3} on 6 questions. While these answers may not be technically accurate, they could still exhibit a moderate level of semantic similarity to the ground truth. Additionally, while we removed the source documents used to generate the synthetic Q\&As, there is no guarantee that these were the only documents containing the necessary information. The model's responses may be based on other retrieved relevant documents that we could not identify and exclude from the prompt.

We attempted to reuse 10\% and 20\% of the training samples (ensuring no overlap with the test set) and removed relevant context from the prompt to create ``Missing Context" training samples, assigning an ``I don't know" (IDK) response as the training label. These samples were then incorporated into the training set, resulting in the models D3-MC10\% and D3-MC20\%. As shown in Table \ref{tab:3}, these models indeed yield lower recall scores on the SynthQA\_MissingContext test set and are more likely to produce IDK-type responses. However, they also tend to generate IDK responses even when source information is available in the prompt, as indicated by the decreased recall scores and increased \#IDK occurrences on the SynthQA test set.

Preventing LLM fine-tuning with RAFT from memorizing and leaking sensitive information remains a challenge. Until a robust solution is developed, we recommend restricting RAFT fine-tuning to non-sensitive data or maintaining separate EDA-LLM assistants, or distinct LoRA adapters, for different user groups.

\section{Conclusion and Discussion}
This paper proposes the use of synthetic question-answer datasets to enhance LLM performance with Retrieval-Augmented Fine-Tuning (RAFT) for EDA tasks. By leveraging broadly available unlabeled documents, we reduce the dependence on scarce labeled data, improving LLM accuracy and functionality in a RAG context. Our exploration into incorporating real user questions as Retrieval-Augmented Few-Shot (RAFS) examples demonstrates further improvements in synthetic data generation, contributing to better model performance. We also address important security concerns by implementing a retrieval-based access control mechanism, ensuring that sensitive information remains restricted to authorized personnel. Furthermore, we assess the potential risks of unintended memorization and data leakage during fine-tuning on synthetic data, providing valuable insights and practical recommendations for mitigating these issues. While the SynthQA test set used for evaluation includes synthetic answers that have not been verified by humans, the results from the Q2A sets demonstrate a clear and promising trend, highlighting the effectiveness of our approach. Future work will focus on utilizing a larger and more diverse set of synthetic QAs to further improve training. Despite the limitations, our results underscore the effectiveness of RAFT with synthetic data in building reliable and secure EDA-LLM assistants, paving the way for future advancements in domain-specific LLM applications.



\bibliographystyle{IEEEtran}
\bibliography{IEEEabrv,main}

\end{document}